\documentclass{llncs}
\usepackage{llncsdoc}

\usepackage[T1]{fontenc}
\usepackage[utf8]{luainputenc}
\usepackage{float}
\usepackage{amsmath}
\usepackage{amssymb}
\usepackage{graphicx}
\usepackage{lmodern}
\usepackage{endnotes}

\usepackage{rotating}
\usepackage{pdflscape}

\let\footnote=\endnote

\begin{document}

%
\title{Bank distress in the news: \\Describing events through deep learning}

\author{Samuel R\"onnqvist\inst{1,2} \and Peter Sarlin\inst{3,4}}
\institute{Turku Centre for Computer Science -- TUCS\\
Department of Information Technologies,\\
\AA bo Akademi University, Turku, Finland\\
\email{sronnqvi@abo.fi}\thanks{Corresponding author}\\
~\\
\and
Applied Computational Linguistics Lab\\
Goethe University Frankfurt am Main, Germany\\
~\\
\and
Department of Economics\\
Hanken School of Economics, Helsinki, Finland\\
~\\
\and
RiskLab Finland\\
Arcada University of Applied Sciences, Helsinki, Finland\\
\email{peter@risklab.fi}
}

\maketitle

\begin{abstract}
While many models are purposed for detecting the occurrence of significant events in financial systems, the task of providing qualitative detail on the developments is not usually as well automated. We present a deep learning approach for detecting relevant discussion in text and extracting natural language descriptions of events. Supervised by only a small set of event information, comprising entity names and dates, the model is leveraged by unsupervised learning of semantic vector representations on extensive text data. We demonstrate applicability to the study of financial risk based on news (6.6M articles), particularly bank distress and government interventions (243 events), where indices can signal the level of bank-stress-related reporting at the entity level, or aggregated at national or European level, while being coupled with explanations. Thus, we exemplify how text, as timely, widely available and descriptive data, can serve as a useful complementary source of information for financial and systemic risk analytics.

\end{abstract}




%

\section{Introduction}
\setlength{\parskip}{0.01em}

Text analytics presents both major opportunities and challenges. On the one hand, text data is rich in information and can be harnessed in traditional ways such as for prediction tasks, while its descriptive depth also supports qualitative and exploratory, yet highly data-driven, analysis. On the other hand, decoding and utilizing the expressive detail of human language is prohibitively difficult. In computational terms, text consists of high-dimensional and often ambiguous symbolic input (words), the semantics of which is a product of complex interactions between parts of the sequences in which they occur (phrases, sentences, paragraphs, etc.). Text is referred to as sparse data due to the high variability relative to number of samples, and unstructured data as the underlying linguistic structure must be inferred from the surface form as part of the analysis process.

We recognize that many applications of text analytics use linguistically rather na\"ive methods, typically operating on a bag-of-words assumption, disregarding word order and operating at the symbolic word-level alone. While these applications generally constitute pioneering work in their respective areas, there is currently ample opportunity for advancement, in particular in the intersection between machine learning, computational linguistics and economics. Following the deep learning paradigm, recent developments in natural language processing \cite{manning2013deeplearning} open up for highly data-driven but linguistically more accommodating analysis methods based on semantic representation learning, which easily can be applied to new domains and tasks.

In this paper, we propose a deep learning setup to address the challenge of building a predictive model able to detect infrequent, coinciding events based on the sparse and unstructured input of text, while leveraging the text data to describe the events as well. Our method includes a heuristic to label text by event information, unsupervised semantic modeling, predictive modeling, aggregation of the prediction signals into indices, and the eventual extraction of descriptions. The approach is to our knowledge novel in providing text descriptions of events defined by non-descriptive data.
We show how it can be applied to the study of risks in the financial system, with relatively little effort required in terms of collecting data for supervision in new tasks, which can be a prohibitive aspect of text analytics.

The study of bank distress is a prime example of a field where the use of text data remains largely uncharted, typically lacking both customized linguistic resources and clear goals for how to best utilize text, which motivates the focus on adaptive methods.
Supervised by only a small set of bank distress events we demonstrate that the method we put forward can provide an index over coinciding stress-related reporting in news over time, which we then use to automatically retrieve descriptions of the events. 
We expect the method accordingly to be applicable to any type of event that recurringly figures in text over time, in connection to specific entities.



In the following section, we discuss previous work related to the problem setting and work that has utilized text data for similar tasks. 
Deep learning background and our setup, including semantic modeling, predictive modeling and evaluation, extraction of descriptions and the related indices are explained in Section 3.
Finally, we report our experiments in Section 4, demonstrating the applicability of our approach to the study of bank stress.

\section{Related work}
The automatic identification of events in chronological text such as news has been explored at least since the 90s, when a DARPA-coordinated effort was organized \cite{allan1998topic} that set the foundation for what is known as \emph{topic detection and tracking} (TDT), where news streams are analyzed in order to identify reporting on new events as well as recurring reporting relating to earlier events. The early detection and tracking methods were data-driven, based on clustering in particular, and intended to capture any kind of event (see, e.g., \cite{yang2000learning}).

A related area of research that since has emerged, mainly stemming from the area of information extraction, is \emph{event extraction}, which aims at extracting complex structured information about events in terms of pre-defined types of events and entities, as well as attributes of events and roles of entities (cf., e.g., \cite{bjorne09sharedtask}).
The event extraction techniques focus on identifying and extracting more specific types of information, with explicit semantic interpretation, in contrast to the TDT approach. As the information of interest is often particular to an expert domain and task, the techniques tend to require substantial expert guidance in terms of designing linguistic patterns or annotating text, which makes them less applicable in new domains where fewer resources may be available to target specific tasks and the information of interest may be difficult to strictly define.
Efforts focusing on the financial domain and identification of specific types of risk include \cite{borsje2010semi,capet2008risk} and \cite{hogenboom2015news}. Tanev et al. \cite{Tanev2008crisis} also explore the combination of data-driven preprocessing with the knowledge-driven approach to extracting events, as they monitor violent and disaster events in news. Hogenboom et al. \cite{Hogenboom2016survey} provide a thorough overview of how event extraction has evolved in various fields.

Parallel to this view on event discovery, which naturally places description of events at its heart, non-text data sources have also been investigated for the detection of significant events, or the risk thereof, such as failure of companies using machine learning \cite{back96bankruptcy,dimitras99failure}. The focus is then primarily on estimating the likelihood that a particular type of event will occur. While the specification of events in text mining tends to be more idiosyncratic to the input data, the events in distress prediction tend to be specified by when they occur and what entities they involve, as is the case in this paper, too. Such event specifications are easier to recombine with new data, including text data given appropriate modeling.

In particular, prediction of bank distress has been a major topic both before and following the global financial crisis. Many efforts are concerned with identifying the
build-up of risk at early stages, often relying upon aggregated
accounting data to measure imbalances (e.g., \cite{cole_predicting_1998,mannasoo_explaining_2009,betz2014predicting}).
Despite their rich information content, accounting data pose major
challenges 
due to restricted access, as well as low reporting frequency and long publication lags. 
A widely available and more	
timely source of information is the use of market data to indicate
imbalances, stress and volatility (e.g., \cite{Groppetal2006,milne2014}).
Yet, market prices provide little or no descriptive information per
se, and only yield information about listed companies or companies'
traded instruments (such as Credit Default Swaps). This points to
the potential value of text as a source for understanding events such as bank distress. 
More generally, central banks are starting to recognize the utility of text data in financial risk analytics, too. \cite{riksbank2015bigdata,boe2015textmining}

The literature on text-based computational methods for measuring risk
or distress is still rather scarce and scattered. For instance, Nyman et al. \cite{Gregoryetal2014}
analyze sentiment trends in news narratives in terms of excitement/anxiety
and find increased consensus to reflect pre-crisis market exuberance,
Soo \cite{soo2013quantifying} analyzes the connection between
sentiment in news and the housing market and Cerchiello et al. \cite{Paola2015} analyse bank risk contagion with both market prices and sentiment index. All three approaches rely on manually-crafted
dictionaries of sentiment-bearing words. While such analysis can provide
interesting insight as early work on processing expressions in
text to study risk, the approach is generally limiting as dictionaries are cumbersome
to adapt to specific tasks, incomplete and unable to handle semantics beyond single words well. Nevertheless, sentiment analysis based on such simple approaches works quite well due to the fact that it relies on human emotions as strong priors in a way that generalizes across tasks and data, and because lower recall may be countered by the scale of the data. 

Malo et al. \cite{malo2014good} explore a linguistically more sophisticated approach that models financial sentiment compositionally, although without semantic generalization, supervised by a custom data set of annotated phrases. Hogenboom et al. \cite{hogenboom2015news} integrate their linguistically aware event extraction techniques with the conventional Value at Risk model to account for certain cases of event-driven market effects.

Data-driven approaches, such as Wang \& Hua \cite{wang2014copula}
predicting volatility of company stocks from earning calls, may avoid
the issues of handcrafted features and manually annotated corpora. Their method, although allegedly providing good predictive
performance gains, offers only limited insight into the risk-related
language of the underlying text data. It also leaves room for further
improvements with regard to the semantic modeling of individual words
and sequences of words, which we address. Further, Lischinsky \cite{lischinsky2011discourse}
performs a crisis-related discourse analysis of corporate annual reports
using standard corpus-linguistic tools, including some data-driven
methods that enable exploration based on a few seed words. His analysis
focuses extensively on individual words and their qualitative interpretation
as part of a crisis discourse, which likewise provides rather limited insight compared to what full sentences are able to communicate. 
Finally, R\"onnqvist \& Sarlin \cite{RonnqvistSarlin2015}
construct network models of bank interrelations based on co-occurrence
in news, and assess the information centrality of individual banks
with regard to the surrounding banking system, a fully data-driven approach that could be further enhanced by semantic modeling and conditioning.

In the following, we introduce the deep learning approach and our particular model, along with further relevant previous work.

\newpage
\section{Methods}

Characterized in part by the deep, many-layered neural networks, a
prevailing idea of the deep learning paradigm is that machine learning
systems can become more accurate and flexible when we allow for abstract
representations of data to be successively learned, rather than handcrafted
through classical feature engineering. By modeling the input data before modeling specific tasks, the networks can learn about regularities in the world and generalize over them, which improves performance on supervised task learning.
For a recent general survey
on deep learning confer Schmidhuber \cite{schmidhuber2015deep}, and
for a more explicit discussion of deep learning in natural language
processing see Socher \& Manning \cite{manning2013deeplearning}. Moreover, Bengio et al. \cite{bengio2013representation} provide a thorough review on the emerging topic of representation learning itself.

While manually designed features help bring structure to the learning
task through the knowledge they encode, they often suffer problems
of being over-specified, incomplete and laborious to develop. Especially
regarding natural language processing, this limits the robustness
of text mining systems and their ability to generalize across languages, domains and
tasks. By exploiting statistical properties of the data,
features can be learned in an unsupervised fashion instead, which
allows for large-scale training not limited by the scarcity of annotated
data. Such intensively data-driven, deep learning approaches have
in recent years led to numerous breakthroughs in application
domains such as computer vision and natural language processing, where
a common theme is the use of unsupervised pre-training to effectively
support supervised learning of deep networks \cite{schmidhuber2015deep}.
We apply the same idea in modeling event-related language in text.

\subsection{Labeling text by event data}
The modeling is founded on connecting two types of data,
text and event data, by entities and chronology. An event data set contains information on dates and names of involved entities, relating to the specific type of event to be modeled. First, a set of regular expression patterns is used to locate the entity names as they occur in the text. Second, an event is associated by the date it occurred and by the relevant timestamp of the document.

In this paper, we focus on news text where publication date is used for matching articles in time, and entity occurrences are indexed at the sentence level. Each sentence $s$ and occurring entity $b$ are cross-referenced against the event data in order to cast the pair as event \emph{coinciding} (1), \emph{non-coinciding} (0) or ambiguous (undefined), according to an inner ($W_{in}$) and outer ($W_{out}$) time window. Formally, the label is defined as:
\[
e_{s,b}=\begin{cases}
1, \text{if } d_{s}-d_{e} \in W_{in}\\
0, \text{if } d_{s}-d_{e} \notin W_{out}
\end{cases}
\]
where for the intervals $W$ holds that $W_{in} \subset W_{out}$. I.e., we label each entity occurrence and its sentence as likely to discuss the event or not likely, whereas uncertain cases that fall outside $W_{in}$ but within $W_{out}$ are not used. Given this heuristic, we effectively expand the data set that is to serve as supervision signal, and the predictive model will learn to generalize across examples and associate relevant language in the text data to the modeled event type.




\subsection{Modeling}

We are interested in modeling the semantics of words and semantic compositionality of sequences of words to obtain suitable representations of the content of the news, to use as features for predicting events and associating text descriptions.
At the word level, distributional semantics exploits the linguistic
property that words of similar meaning tend to occur in similar contexts
\cite{harris1954distributional}. Word contexts are modeled to yield
distributed representations of word semantics as vectors, as opposed to declarative formats, which allow
measuring of semantic similarities and detecting analogies without
supervision, given substantial amounts of text \cite{Schutze:1992:DM:147877.148132,schutzePedersen1995irwordsenses,mikolov2013efficient}.
The distributional semantic modeling captures the nature of words in a broader sense, in the directions of syntax and pragmatics.
These word vectors provide an embedding into a continuous semantic space where
the symbolic input of words can be geometrically related to each other,
thus supporting both the predictive modeling in this paper and a multitude
of other natural language processing tasks (e.g., tagging, parsing
and relation extraction) \cite{manning2013deeplearning,bengio2013representation}.

While traditionally modeled by counting of context words, predictive
models have eventually taken the lead in terms of performance
\cite{baroni2014don}. Neural network language models in particular
have proved useful for semantic modeling, and are especially practical
to incorporate into deep learning setups due to their dense vectors
and the unified neural framework for learning. Mikolov et al. \cite{mikolov2013efficient}
have put forward an efficient neural method that can learn highly
accurate word vectors as it can train on massive data sets in practical
time (a billion words in the order of a day on standard architecture).

\begin{figure}[t]
\begin{center}
\includegraphics[width=0.7\columnwidth]{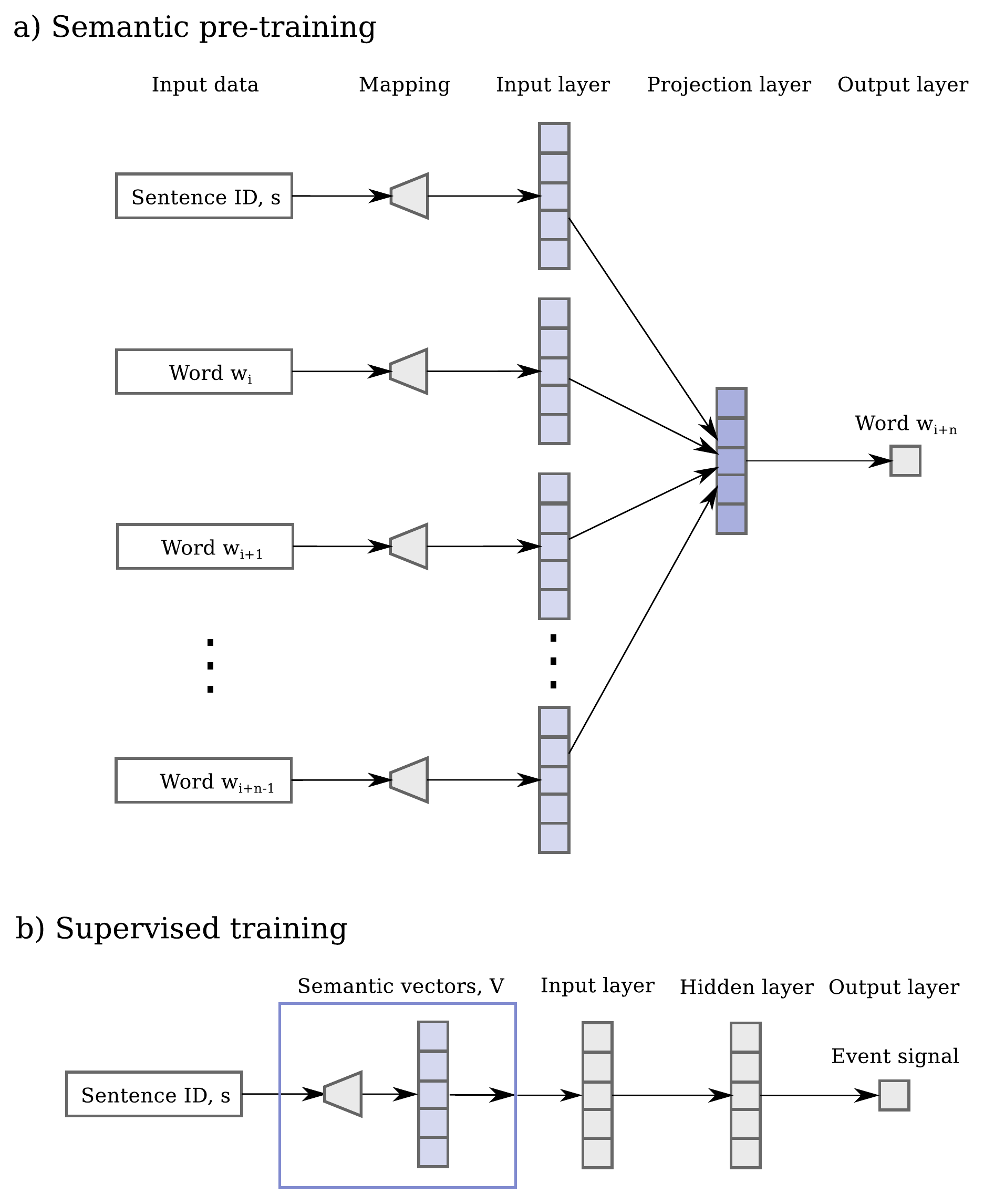}
\end{center}
\protect\caption{Deep neural network setup for (a) pre-training of semantic vectors, and
(b) supervised training against event signal $e$.}
\label{fig_deep_setup}
\end{figure}

Subsequently, Le \& Mikolov \cite{le2014distributed} extended the
model in order to represent compositional semantics (cf. \cite{mitchell2010composition})
of sequences of words, from sentences to the length of documents,
which they demonstrated to provide state-of-the-art performance on
sentiment analysis of movie reviews. Methods based on other neural architectures and explicit sentence structure have since gained slightly improved performance \cite{irsoy2014deep,tai2015improved}, but require parse trees as pre-structured input and are therefore not as flexible. Analogous to the sentiment analysis task, we employ
the distributed memory method of Le \& Mikolov to learn vectors for sentences in news
articles, where entities are mentioned, and use them for learning to predict the probability of an event. Hence, when providing bank distress events, the task can be understood as a type of risk sentiment analysis that models language specific to the type of event, rather than more general expression of emotions explicitly.

Text sequences are also commonly modeled by recurrent neural networks such as Long Short-Term Memory (LSTM) networks \cite{hochreiter1997long}, but these are not as efficient as feed-forward topologies with fixed context size in terms of speed.
The sentence vector we use is a practical fixed-size
representation suitable as input to a feed-forward network. 
The input sequence of words may have a vocabulary size in the order of a million words,
but the sentence vector represents the necessary semantics of each sentence as a single dense vector with a dimensionality of typically 50-1000.
The reduction from sparse sequence to a fixed-length, dense representation 
helps train the predictive model against a signal
corresponding to a comparatively tiny number of events.

Our deep neural network for predicting events from text, outlined
in Fig.~\ref{fig_deep_setup}, is trained in two steps: through learning
of sentence vectors as pre-training (\ref{fig_deep_setup}a), followed
by supervised learning against the event signal $e$ (\ref{fig_deep_setup}b).
The use of the distributed memory model of \cite{le2014distributed} in the first step is explained
in the following.

The modeling of word-level semantics works by running a sliding window over text, taking a sequence of
words as input and learning to predict the next word (e.g., the 8\textsuperscript{th}
in a sequence), using a feed-forward topology where a projection layer
in the middle provides the semantic vectors once the connection weights have been
learned. A semantic vector $V_i$ is the fixed-length, real-valued pattern of activations reaching the projection layer for network input $i$. The projection layer provides a linear combination that enables
efficient training on large data sets, which is important in achieving
accurate semantic vectors. In addition, the procedure of \cite{le2014distributed} for sentence vector training includes the
sentence ID as input, functioning as a memory for the model that allows
the vector to capture the semantics of continuous sequences rather
than only single words; the sentence ID in fact can be thought of as
an extra word representing the sentence as global context and informing the prediction
of the next word. While the prediction from word context to word constitutes a basic neural language model, the sentence ID conditions the model on the sentence and forces the sentence vector to capture the semantics that is particular to the sentence rather than the language overall. Formally, the pre-training step seeks to maximize
the average log probability:
\[
\frac{1}{k-n}\sum_{i=1}^{k-n}\mathrm{log}\, p(w_{i+n}|s,w_{i},...,w_{i+n-1})
\]
over the sequence of training words $w_{1},w_{2},...,w_{k}$ in sentence
$s$ with word context size $n$. In the neural network, an efficient
binary Huffman tree coding is used to map sentence IDs and words to activation patterns in the
input layer and the hierarchical softmax output layer (by referencing vectors of a matrix $D$), which imposes a basic organization of words by frequency. 
The projection layer output is a function of the average of sentence vector $V_{s}$ and word vectors of the context $\{V_{w_{j}}|j\in [i,i+n[\}$, 
which means that a single vector can easily be extracted once the model is trained. The sentence vector is extractable as $V_{s}=\beta+UD_{s}$, where $U$ is the learned projection layer weight matrix and $\beta$ is the bias parameter.

The second modeling step (Fig. \ref{fig_deep_setup}b) is a normal
feed-forward network fed by the sentence vectors $V_{s}$ (pertaining
to the set of sentences $S$), which we train by Nesterov's Accelerated Gradient \cite{nesterov1983nag} and backpropagation \cite{rumelhart1986learning} to predict
distress events $e\in\left\{ 0,1\right\}$. Hence, the objective is to maximize the average
log probability:
\[
\frac{1}{|S|}\sum_{s\in S}\mathrm{log}\, p(e_{s}|V_{s})
\]

The network has two output nodes for $e\in\{0,1\}$ in a softmax layer that applies a cross-entropy loss function. In the trained network, the posterior probability $M(V_{s})=p(e_{s}=1|V_{s})$ reflects the relevance of sentence $s$ to the modeled event type and is derived by:
\[
p(e_{s}=j|V_{s}) = \frac{e^{y_{j}}}{e^{y_{0}}+e^{y_{1}}};\;y=\sigma(\beta^{2}+U^{2}\sigma(\beta^{1}+U^{1}V_{s}))
\]
where $\sigma$ can be any non-linear activation function (e.g., sigmoid, hyperbolic tangent or rectified linear) and $U$ are again the learned weight matrices. 

In the following sections, we discuss how the model is used for classification and evaluated by its classification performance, as we apply a threshold on the model output $M(V_{s})$, as well as on aggregate functions of it.



\subsection{Evaluation and aggregation}

Assuming that the distribution of events for a particular entity is sparse over time, the procedure for matching events to text produces examples with skewed class frequencies. Moreover, it is likely that the user has an imbalanced preference between types of errors, preferring a sensitive system to detect possible events and provide means for further investigation in the form of descriptions, rather than missing an event. This requires extra care in evaluation.

We evaluate the performance of the predictive model to guide hyperparameter optimization and asses the quality of indices that it will produce, and importantly to provide a quantitative quality assurance for the information content of the descriptions
we extract. We use the relative Usefulness measure ($U_{r}$) by Sarlin
\cite{Sarlin2013b}, as it is commonly used in distress prediction
and intuitively incorporates both error type preference ($\mu$) and
relative performance gain of the model over consistently choosing the majority class. Based on the combination of
negative/positive observations ($\mathrm{obs}\in\left\{0,1\right\}$) and negative/positive
predictions ($\mathrm{pred}\in\left\{0,1\right\}$), we obtain the cases of true
negative ($TN\equiv obs=0\wedge pred=0$), false negative ($FN\equiv obs=1\wedge pred=0$),
false positive ($FP\equiv obs=0\wedge pred=1$) and true positive
($TP\equiv obs=1\wedge pred=1$), for which we can estimate probabilities
when evaluating our predictive model. Further, we define the baseline
loss $L_{b}$ to be the best guess according to prior probabilities
$p(\mathrm{obs})$ and error preferences $\mu$ (Eq. 1) and the model
loss $L_{m}$ (Eq. 2):

\begin{equation}
L_{b}=\mbox{min}\begin{cases}
\mu\cdot p(\mathrm{obs}=1)\\
(1-\mu)\cdot p(\mathrm{obs}=0)
\end{cases}
\end{equation}

\begin{equation}
L_{m}=\mu\cdot p(FN)+(1-\mu)\cdot p(FP)
\end{equation}

From the loss functions we derive Usefulness in absolute ($U_{a}$) and relative terms ($U_{r}$):

\begin{equation}
U_{r}=\frac{U_{a}}{L_{b}}=\frac{L_{b}-L_{m}}{L_{b}}
\end{equation}

While absolute Usefulness $U_{a}$ measures the gain vis-\`a-vis
the baseline case, relative Usefulness $U_{r}$ relates gain to
that of a perfect model (i.e., Eq. 5 with $L_{m}=0\Rightarrow U_{a}=L_{b}$).
Usefulness functions both as a proxy for benchmarking the model (testing)
and to optimize its hyperparameters (validation). Usefulness can also be related to the in text mining
widely used $F$-score\cite{vanRijsbergen1979} (based on $\mathrm{precision}=p(\mathrm{obs}=1|\mathrm{pred}=1)$
and $\mathrm{recall}=p(\mathrm{pred}=1|\mathrm{obs}=1)$):

\begin{equation}
F_{\beta}=(1+\beta^{2})\cdot\frac{\mbox{precision\ensuremath{\cdot}recall}}{(\beta^{2}\cdot\mbox{precision})+\mbox{recall}}
\end{equation}
which similarly can account for varying preferences by
its $\beta$ parameter, although not gain. The $F_{\beta}$-score
assigns $\beta$ times as much importance to recall as to precision
(i.e., preference for completeness over exactness)\cite{vanRijsbergen1979},
which is analogous to but not directly transferable to the $\mu$
parameter in the Usefulness measure. While the $F$-score is commonly
seen to maximize completeness versus exactness of true positives,
the parameter can also be seen as a priority to minimize false negatives
versus false positives (FN prioritized over FP when $\beta>1$). As
a heuristic, we map the balanced, standard $F_{1}$-score with $\beta=1$
to $U_{r}$ with $\mu=0.5$, and match deviations from these preferences
according to $\beta=\mu/(1-\mu)$.

In order to influence the sensitivity of the model, we may classify a sentence by a threshold on the positive-class posterior probability: 
\[
p(e_{s}=1|V_{s})\geq t
\]
The threshold is optimized on the validation set with respect to Usefulness at a given preference, and applied to the test set for evaluation. 

However, evaluating classification at an aggregated entity level rather than the level of sentence instances is more suitable to the use case, and likely more robust as the classification then combines evidence from multiple observed occurrences in the text. Instead of the direct posterior probability, at the entity level we classify by the index defined in Eq. 5 below;	 i.e., an event is signaled for the entity if: 
\[
I(p,b)\geq t
\]

Furthermore, evaluation on the sentence vector level with a randomized set split into train, validation and test set may produce somewhat optimistic results, as specific language related to one particular event can be expected to be shared among several instances. Thus, the evaluation would not truly reflect how well the model can be expected to generalize across events of the same type, including future occurrences. To counter the bias, we sample the cross-validation folds according to a \emph{leave-N-entities-out} strategy (or leave-N-banks-out), based on entity rather than sentence instance, such that discussion about a particular entity is compartmentalized into a single set. In case of very frequent entities that would cause very skewed fold sizes, the instances may be split by period such that the more recent occurrences are placed in the latter set (e.g., test rather than validation set) to minimize possible cross-contamination.

\subsection{Event indices}

By aggregating posterior probabilities we form an index to reflect the level of event-related reporting about an entity over time, thereby guiding exploration and extraction of descriptions, while it also serves as the signal that we evaluate against. The entity-level relevance index $I:p\times b\rightarrow[0,1]$ is formalized as:

\begin{equation}
I(p,b)=\frac{1}{|S_{p,b}|}\sum_{s\in S_{p,b}}M(V_{s})
\end{equation}
over the sentences $S_{p,b}$ that mention entity $b$ in period $p$,
where $M(V_{s})=p(e_{s}=1|V_{s})$ gives the posterior probability of the trained neural network model.

In order to obtain better overview, it is motivated to further group entities and aggregate their indices. In the experiments, we first aggregate from sentences to banks, and then from banks to countries to highlight national differences across Europe. The second-level index (or country-level index) is a weighted average, defined as:

\begin{equation}
I'(p,c)=\frac{1}{|B_{c}|}\sum_{b\in B_{c}}I(p,b)\cdot|S_{p,b}|
\end{equation}
where $B_{c}$ is the set of entities in category/country $c$.
Finally, we define a top-level index that summarizes the level of relevant reporting for all modeled entities as a global average of vectors:

\begin{equation}
I''(p)=\frac{1}{|S_{p}|}\sum_{s\in S_{p}}M(V_{s})
\end{equation}
where $S_{p}$ is the set of vectors for all entity-mentioning sentences in period $p$.


\subsection{Extraction of descriptions}

As the neural network in the second step of the setup has been trained and the hyperparameters
optimized by cross-validation, it can be applied to sentence vectors $V$ in order to use the posterior
probability $M(V)$ as a relevance score with respect to the event type. The indices (Eq. 5, 6 and 7) provide overview over time and can highlight peaks and periods with elevated volumes of event-related discussion, which can be more closely investigated by retrieving descriptions of the underlying events.

Given a specific period and entity or set of entities, the basic principle in retrieving descriptions is to filter and rank pieces of text based on the posterior probability of the predictive model for the corresponding semantic vector. In the current setup, we perform the semantic modeling on the sentence level, which simplifies the process of retrieving relevant and specific passages. The semantic modeling can be applied to any type of textual unit, including complete documents, but that requires additional measures for locating the interesting parts within the broader context. R\"onnqvist and Sarlin \cite{ronnqvist2015detect} explore this by similarly training a predictive model on document vectors and successfully applying it on word vectors, to weight the relevance individual words within the context. In current experiments, we find that, while their method works for document vectors that are trained on a larger number of words per vector, it does not work as well for sentence vectors, as they tend to be less similar to the word vectors of the same model. Overall, the extracts as presented in Section 4 are qualitatively better when produced based on sentence vectors.

Vectors are trained only for sentences that mention target entity names, as it would be infeasible in terms of memory to model each sentence separately for a large corpus, and because the direct discussion about the entities is of primary interest. The near context of such sentences however tend to support interpretation and are useful to include in presentation. The semantic model supports inference of vectors for at train-time unseen sentences, although with noisier results. We infer vectors and predict the relevance of the sentences immediately before and after sentences in which entities occur, as there is strong dependency between neighboring sentences and a combined score of the expanded context may produce more robust predictions. The combined score for an excerpt is calculated as:
\begin{equation}
x_{i}=max\begin{cases}
M\left(V_{S_{i}}\right)\\
M\left(\frac{1}{n}\sum_{j=1}^{n}V'_{S_{i-1}}\right)\\
M\left(\frac{1}{n}\sum_{j=1}^{n}V'_{S_{i+1}}\right)
\end{cases}
\end{equation}
which includes one sentence before and after sentence $S_{i}$. $V'$ is a stochastic, inferred vector and $n$ is the number of samples (e.g. 100).

The excerpts are ranked according to the score for presentation and offer a preview of the most prominent event-related discussion, which may be retrieved in full from the individual articles. The experiments that follow demonstrate the utility of the excerpts in highlighting the specific forces that drives the index, as we apply the method to model bank distress.

\section{Experiments}

We test the deep neural network setup for modeling event-related language on European bank distress events and news data, in order to demonstrate the value it can bring in helping to identify and understand past, ongoing or mounting events. In the following, we discuss the data we use, the modeling in practice, and our quantitative evaluation results. Finally, we provide a qualitative analysis of the indices and related events by means of their associated descriptions, going from the general, higher-level view to the more specific.


\subsection{Data}

The event data set for this study
covers data on large European banks as entities, spanning periods before,
during and after the global financial crisis of 2007--2009. We include
101 banks for which 243 distress events have been observed during 2007Q3--2012Q2.
Following Betz et al. \cite{betz2014predicting}, the events include
government interventions and state aid, as well as direct failures
and distressed mergers. In addition, we map each bank to the country or countries where it is registered, to allow for aggregation of results to the country level.

The text data consist of news articles from Reuters online archive
from the years 2007 to 2014 (Q3). The data set includes 6.6M articles
(3.4B words). 
Bank name occurrences are located using a set of regular expressions that cover common spelling variations and abbreviations.
The patterns have been iteratively developed against the data to increase
accuracy, with the priority of avoiding false positives (in accordance
to \cite{RonnqvistSarlin2015}). Scanning the corpus, 262k articles
and 716k sentences are found to mention any of the 101 target banks.

We set the inner time window from 8 days before to 45 days after the event ($W_{in}=[-8, 45]$), and the outer from 120 days before to 120 days after ($W_{out}=[-120, 120]$), as optimized through the evaluation scheme discussed in
Section 3.3. In total, 386k sentences are successfully labled and used for training and evaluation, as they fall within the span of the event data and are not deemed ambiguous cases. As expected, the class distribution is highly skewed, with 9.0\% of the 386k cases being labeled as coinciding.



\subsection{Semantic pre-training}

First, the semantic pre-training step is performed to obtain sentence vectors for each of the 716k sentences, to be used both for training, evaluation and deployment of the model. In order to improve the word representations of the model, by extending the data coverage and letting them capture the semantics of both general English in news reporting as well as bank-specific language, the rest of the corpus is also sampled. This is achieved by running the model without the sentence-ID-related component for sentences without bank occurrences. The whole training process is repeated in multiple iterations with decreasing learning rate. We optimized the sentence vector length to 600 and context size to 5 by cross-validation. We also tested the influence of text sequence lengths, and found that training a vector on multiple sentences achieved slightly worse predictive performance, while vectors trained at sentence and document level were comparable.

\subsection{Predictive modeling and evaluation}

Following the semantic pre-training, we train a predictive neural
network model with 3 layers. The input layer has 600 nodes, corresponding
to the semantic vectors, and the output layer has two nodes corresponding to distress/tranquil states.
A set of tuples including sentence vectors $V_{s}$, entity $b$ and labels $e_{s,b}$ are complied as data for modeling.

\setlength{\tabcolsep}{0.32cm}
\begin{table}
\centering
\begin{tabular}{ccccccccc}
 & \multicolumn{4}{|c|}{Random sampling} & \multicolumn{2}{c|}{Leave-N-banks-out}\tabularnewline
 & \multicolumn{2}{|c}{Vector-level} & \multicolumn{2}{c|}{Aggregated} & \multicolumn{2}{c|}{Vector-level}\tabularnewline

$\mu$ & $\bar{U}_{r}(\mu)$ & $\sigma_{U}$ & $\bar{U}_{r}(\mu)$ & $\sigma_{U}$ & $\bar{U}_{r}(\mu)$ & $\sigma_{U}$\tabularnewline
\hline 
\hline 
0.1 & -0.004 & 0.004 & -0.022 & 0.029 & -0.013 & 0.013\tabularnewline
0.3 & -0.007 & 0.004 & -0.015 & 0.013 & -0.032 & 0.026\tabularnewline
0.5 & 0.002 & 0.005 & -0.014 & 0.010 & -0.039 & 0.036\tabularnewline
0.6 & 0.013 & 0.007 & -0.015 & 0.012 & -0.038 & 0.039\tabularnewline
0.7 & 0.038 & 0.011 & 0.027 & 0.030 & -0.026 & 0.029\tabularnewline
0.8 & 0.095 & 0.019 & 0.156 & 0.029 & -0.008 & 0.044\tabularnewline
0.85 & 0.157 & 0.026 & 0.260 & 0.030 & 0.025 & 0.048\tabularnewline
0.875 & 0.207 & 0.028 & \textbf{0.326} & 0.030 & 0.039 & 0.133\tabularnewline
0.9 & \textbf{0.275} & 0.054 & 0.268 & 0.031 & \textbf{0.083} & 0.114\tabularnewline
0.925 & 0.253 & 0.041 & 0.148 & 0.040 & 0.040 & 0.109\tabularnewline
0.95 & 0.106 & 0.044 & -0.009 & 0.038 & -0.052 & 0.153\tabularnewline
\tabularnewline
\end{tabular}
\\
\protect\caption{Cross-validated predictive performance as relative Usefulness over
preferences between types of error ($\mu$), evaluated at vector and aggregated bank level with random sampling, and at vector level with leave-N-banks-out sampling.}
\label{eval_table1}
\end{table}

\setlength{\tabcolsep}{0.32cm}
\begin{table}
\centering
\begin{tabular}{ccccccccc}
 & \multicolumn{8}{|c|}{Leave-N-banks-out, aggregated}\tabularnewline
$\mu$ & $\bar{U}_{r}(\mu)$ & $\sigma_{U}$ & $\bar{F}_{\beta}$ & $\sigma_{F}$ & $\bar{T\! N}$ & $\bar{F\! N}$ & $\bar{F\! P}$ & $\bar{T\! P}$\tabularnewline
\hline 
\hline 
0.1 & -0.014 & 0.042 & 0.497 & 0.000 & 516 & 68 & 0 & 0\tabularnewline
0.3 & -0.011 & 0.022 & 0.087 & 0.015 & 516 & 68 & 0 & 0\tabularnewline
0.5 & -0.015 & 0.029 & 0.031 & 0.013 & 516 & 68 & 0 & 0\tabularnewline
0.6 & -0.013 & 0.027 & 0.032 & 0.020 & 515 & 68 & 1 & 0\tabularnewline
0.7 & -0.003 & 0.038 & 0.087 & 0.063 & 511 & 65 & 4 & 3\tabularnewline
0.8 & 0.048 & 0.154 & 0.314 & 0.171 & 472 & 53 & 44 & 15\tabularnewline
0.85 & 0.122 & 0.147 & 0.434 & 0.153 & 435 & 45 & 80 & 22\tabularnewline
0.875 & \textbf{0.123} & 0.173 & 0.529 & 0.174 & 374 & 38 & 142 & 30\tabularnewline
0.9 & 0.081 & 0.162 & 0.629 & 0.189 & 308 & 31 & 208 & 37\tabularnewline
0.925 & -0.006 & 0.173 & 0.741 & 0.190 & 151 & 14 & 364 & 54\tabularnewline
0.95 & -0.075 & 0.160 & 0.901 & 0.125 & 38 & 4 & 477 & 64\tabularnewline
\tabularnewline
\end{tabular}
\\
\protect\caption{Cross-validated predictive performance as relative Usefulness and $F$-score over
preferences between types of error ($\mu$) and recall/precision ($\beta$), evaluated at bank level with leave-N-banks-out sampling. Mean confusion matrix values are included, too.}
\label{eval_table2}
\end{table}

We evaluate the predictive model with the four combinations of sampling method and level of evaluation, discussed in Section 3.3. The baseline evaluation with random sampling at the level of sentence vectors is reported in Table \ref{eval_table1} (left), providing 27.5\% relative Usefulness, i.e., performing significantly better than majority class prediction even with the highly skewed class distributions. By comparison, evaluation at the aggregated bank level (classifying by $I(p,b)$ (Eq. 5) rather than $M(V)$) reduces noise from single sentences and stabilizes prediction, thereby increasing performance to 32.6\% (Table \ref{eval_table1}, center). These results show that the model is effective in linking the relevant pieces of text to the bank distress events, hence, providing a first assurance of the quality of the descriptions we will retrieve. Further, we evaluate based on leave-N-banks-out sampling, i.e., the cross-validation folds of vectors are organized by bank, such that the vectors of banks used for testing are held out of training. While this produces lower Usefulness scores, it is a more realistic estimate of future performance in the context of deploying the model on unseen banks or future data. With vector-level evaluation we reach 8.3\% relative Usefulness (Table \ref{eval_table1}, right), while bank-level aggregation again stabilizes prediction and improves performance to 12.3\% of available Usefulness (Table \ref{eval_table2}).

We find the optimal network (50 rectified linear hidden
nodes), hyperparameters for the NAG training algorithm
to train its weights, and threshold on $M(V)$ or $I(b,p)$ for classifying $e\in\left\{ 0,1\right\}$, after which we evaluate performance by $U_r$ of the optimal model. We trained the network by randomized
5-fold cross validation with one fold for validation and one for
testing, in multiple reshuffles of the data set. The evaluation yielded an area under the ROC
curve of 0.712 with a standard deviation $\sigma=0.008$ with random sampling evaluated at vector level, and an area of 0.645 ($\sigma=0.083$) with leave-N-banks-out sampling evaluated at the aggregated bank level.

Following previous studies \cite{betz2014predicting,Peltonenetal2015}, we make
use of a skewed preference $\mu\approx0.9$ (i.e., missing a crisis is about 9
times worse than falsely signaling one). From the viewpoint of policy,
highly skewed preferences are particularly motivated when a signal
leads to an internal investigation, and reputation loss or other political
effects of false alarms need not be accounted for. 
While our model is not robust to low levels of $\mu$, we can see
in Table \ref{eval_table2} that Usefulness is positive and peaking as $\mu$
nears 0.9. Meanwhile, $F$-score is reaching its maximum at the extreme preference, which is an indication of its failure to capture gain over the majority class baseline.

We conclude that
at $\mu=0.9$ with vector-level evaluation and at $\mu=0.875$ with aggregated evaluation the model has decent predictive performance by capturing up to
33\% of available Usefulness and 12\% in the more conservative leave-N-banks-out sampled exercise. 
To relate the results we may confer Betz et al. \cite{betz2014predicting} who obtain $U_{r}$ of 19-42\% and Peltonen et al. \cite{Peltonenetal2015} with 58-64\%.
The latter incorporates network linkages, which we currently do not model, although this is possible to extract from text as well (cf. \cite{RonnqvistSarlin2015}).
In both cases they test a selection of models to predict bank distress using conventional data sources. These are the most similar experiments available, although not necessarily strictly comparable. A direct comparison of usefulness is in principle impossible as different data and prediction tasks will yield different results, such as the broader sample and earlier forecast horizons in Betz et al. \cite{betz2014predicting}. Nevertheless, our evaluation results show that we are able to extract a stress signal from text alone. While it does not surpass the performance achieved for other tasks and samples, it does achieve acceptable levels and provides a quantitative quality assurance of the text extracts. The results also point toward the likely benefit of incorporating both text and conventional data in bank distress prediction.

\subsection{A descriptive stress index for Europe}

Having trained the network and evaluated its predictive performance,
we can reliably extract indices of stress at the different levels of aggregation together with extracts to describe them.
In this section, we discuss patterns recognizable in the top-level view, with a summary of what we are able to learn from the associated descriptions. 
The following sections continue with a breakdown into countries and banks, which supports a more targeted qualitative analysis.

\begin{landscape}
\begin{figure*}[p]
\includegraphics[width=1.6\textwidth]{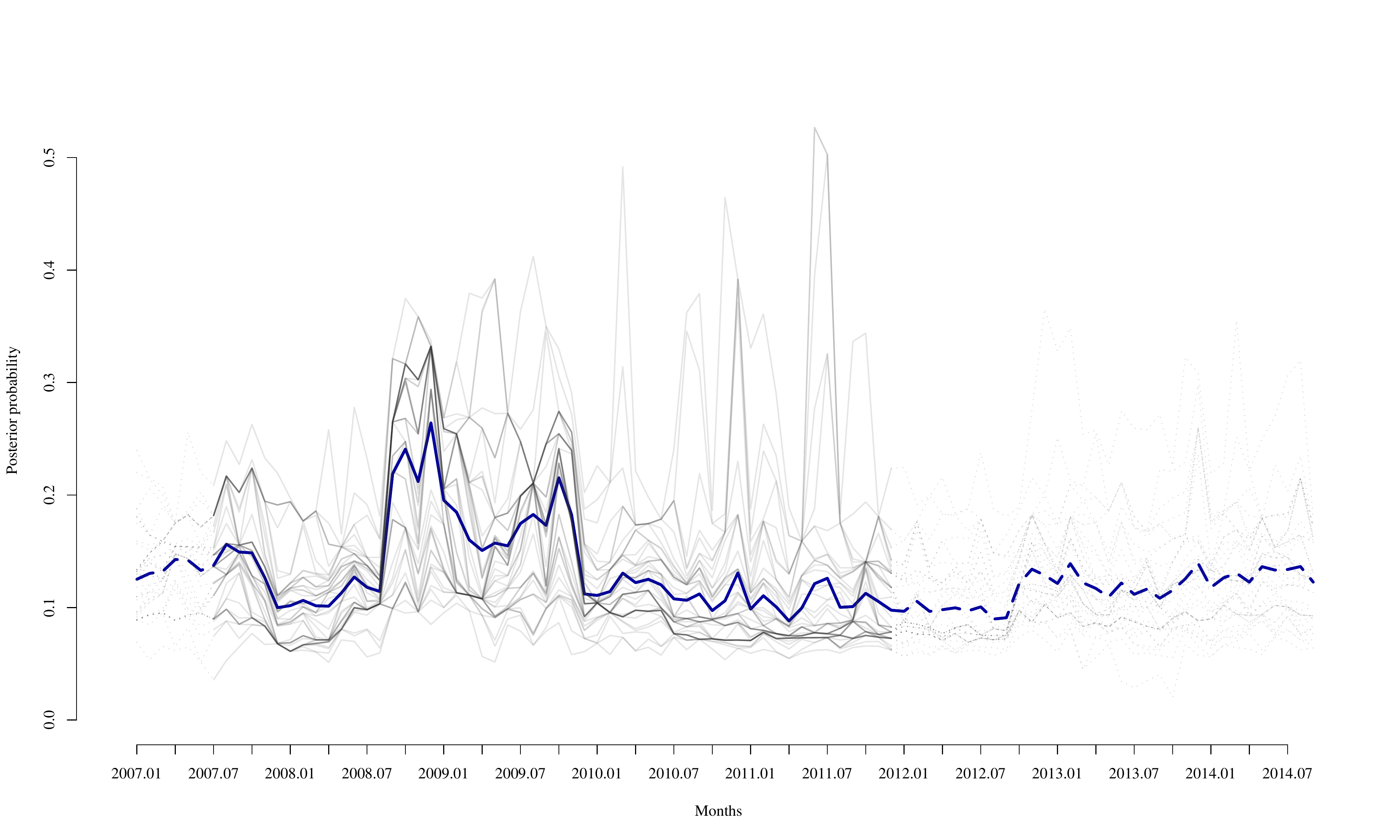}

\protect\caption{Raw distress reporting. Distribution of posterior probabilities over time for sentence vectors, indicating the levels of news reporting relating to bank stress. The blue line indicates
mean, faded lines every 2nd percentile, and dotted lines predictions outside the event sample.}

\label{fig_raw}
\end{figure*}
\end{landscape}

First, Fig. \ref{fig_raw} provides an overview of the raw distress reporting in Europe over the recent years, in terms of distributions of posterior probabilities of the sentence vectors, illustrated through their percentiles.
The time span July 2007 to June 2012 is covered by the event data, and the rest is produced by applying the trained model.
This distribution communicates the dynamics of the stress situation in Europe, while the mean (index $I''$ of Eq. 7) summarizes the general trends.

The index shows a sharp double peak starting September 2008, which coincides with the outbreak of the financial crisis. Prior to the most significant peaks, one can also observe elevated values between August and October 2007, pointing to early discussion on the significance of subprime activities overall and liquidity in European banks. The outbreak of the financial crisis in 2008 is followed by over a year of relatively high stress, where a substantial part of the cross section is elevated. A second significant and similar peak of the stress index is reached in October 2009. At the end of 2010 and 2011, one can observe notable jumps in the most extreme percentiles, whereas the rest of the cross section remains largely unaffected.

At a general level, we observe that the peak in September 2008 relates to overall distress in financial markets due to the collapse of Lehman Brothers in mid-September. However, the fact that values at the top of the distribution appear rather unstable from month to month reflects that different banks are being mentioned over time and usually not persistently across months in distress contexts. By observing increases and peaks in the index of an individual bank or banks in a country, we can identify specific events of possible relevance to distress. 

\subsection{Country-level stress, descriptions and interpretation}

\begin{figure*}
\includegraphics[width=0.99\textwidth]{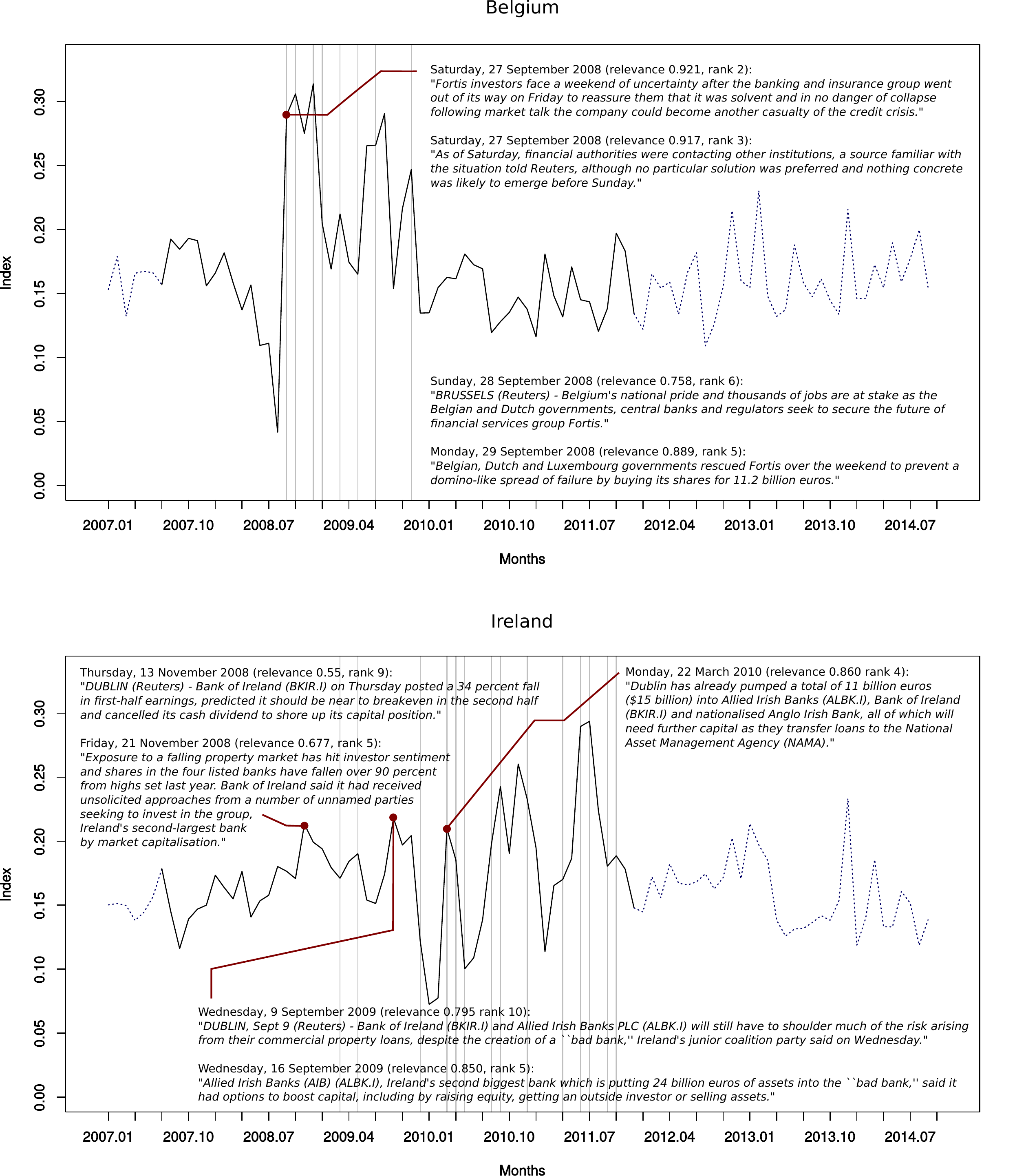}
\protect\caption{Distress index for Belgium and Ireland, with key periods marked and informative excerpts selected from the top-10 of each period and country. Vertical lines indicate distress events and dotted lines out-of-sample predictions. Quotes are from Reuters at given dates.}

\label{fig_excerpts1}
\end{figure*}

\begin{figure*}
\includegraphics[width=0.99\textwidth]{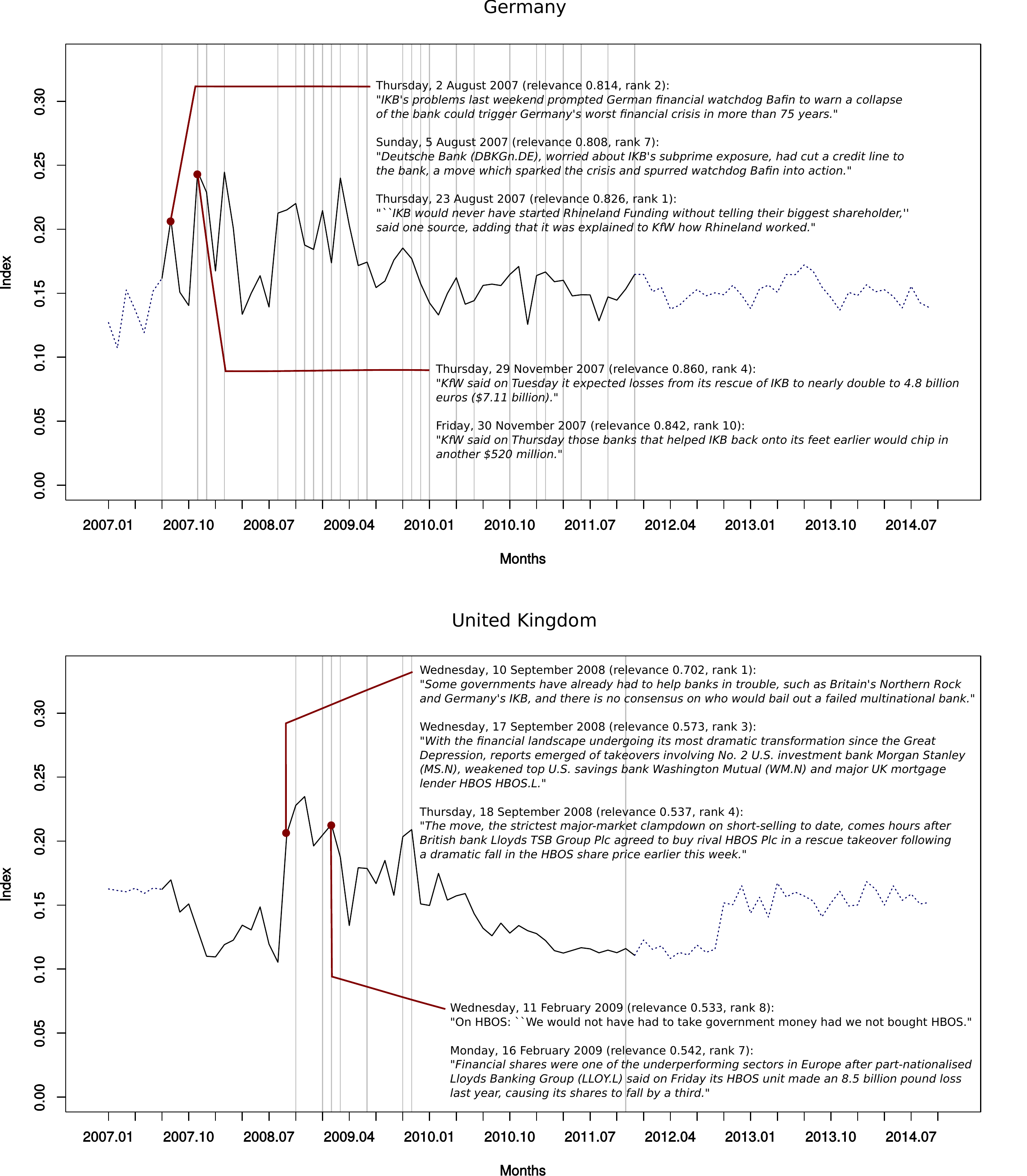}
\protect\caption{Distress index for Germany and the United Kingdom, with key periods marked and informative excerpts selected from the top-10 of each period and country. Vertical lines indicate distress events and dotted lines out-of-sample predictions. Quotes are from Reuters at given dates.}

\label{fig_excerpts2}
\end{figure*}

From the general stress index for Europe, this section moves to a more granular perspective on stress, closer to the level of the events being modeled. We measure stress-related discourse for countries for a more targeted stress measure, which also allows for more economic interpretation of developments, as we study the top-ranking excerpts at key points. Thus, we now aggregate posterior probabilities over time for sentence vectors, indicating the levels of news reporting relating to bank stress, but selectively at a country level (according to Eq. 6). Fig. \ref{fig_excerpts1} shows the developments in stress-related discussion for Belgium and Ireland and Fig. \ref{fig_excerpts2} for Germany and UK. The figures illustrate stress levels as time series, as well as they annotate peaks of distress levels with top-ranked excerpts. In the appendix, we include plots in Figs. \ref{fig_all_index1} and \ref{fig_all_index2} for the other countries whose banks we model.

In Fig. \ref{fig_excerpts1}, the stress levels for Belgium peak in September 2008. Looking at top-ranked excerpts, September 27 is coupled with a range of rumours in media, yet no official release or actions to mitigate the weakened position of particularly Fortis Bank. Then, the next days we see a bailout of Fortis being discussed as the Belgian, Dutch and Luxembourg governments rescued Fortis. Likewise, the lower chart for Ireland in Fig. \ref{fig_excerpts1} shows increased concerns over Bank of Ireland and other large Irish banks in November 2008, as both their earnings and shares were significantly falling. After a range of actions by the state, distress levels were still peaking in September 2009, which is particularly related to the amounts that Allied Irish Banks was putting into the Irish "bad bank". Still, in March 2010 three large Irish banks were still transferring large loans to the National Asset Management Agency (NAMA). Thereafter the most acute stress decreased and has since been at lower levels, although remaining somewhat volatile.

Fig. \ref{fig_excerpts2} provides similar stress time series and top-ranked excerpts, but for Germany and the UK. Germany can be seen to signal already in August 2007, when IKB's problems were highlighted to potentially lead to "Germany's worst financial crisis in more than 75 years". Three days after this news, Deutsche Bank cut a credit line to IKB, as they were worried about IKB's subprime exposures, which further triggered distress in the German banking sector. One reason to the failure of IKB related to an offshore portfolio that was kept off IKB's balance sheet by Rhineland Funding, which is said to have been explained to the largest shareholder KfW. The same large shareholder is then a few months later involved in helping IKB back on its feet with a hefty 4.8 billion euros, as well as additional smaller support afterwards. For the UK, stress increased in September 2008, relating not only to previous aid to the UK-based Northern Rock but also to Germany's IKB. Here, we see an example of cross-border, systemic effects of bank distress. Only a few days later in conjunction with a strict clampdown on short-selling, UK-based bank Lloyds Group bought rival HBOS in a rescue takeover. Ironically, a few months later in February 2009 Lloyds in partly nationalized as its HBOS unit made an 8.5 billion pounds loss the year before.

\subsection{The case of Fortis and IKB Bank}

This section takes a final step towards more granular output by providing a stress measure for individual banks (according to Eq. 5). As with the country-level aggregates, we can aggregate posterior probabilities for sentence vectors selectively by bank. This output could be derived for each of the 101 banks, although here we focus on the stress reporting for two banks, namely Fortis and IKB Bank. 

\begin{figure*}
\includegraphics[width=0.99\textwidth]{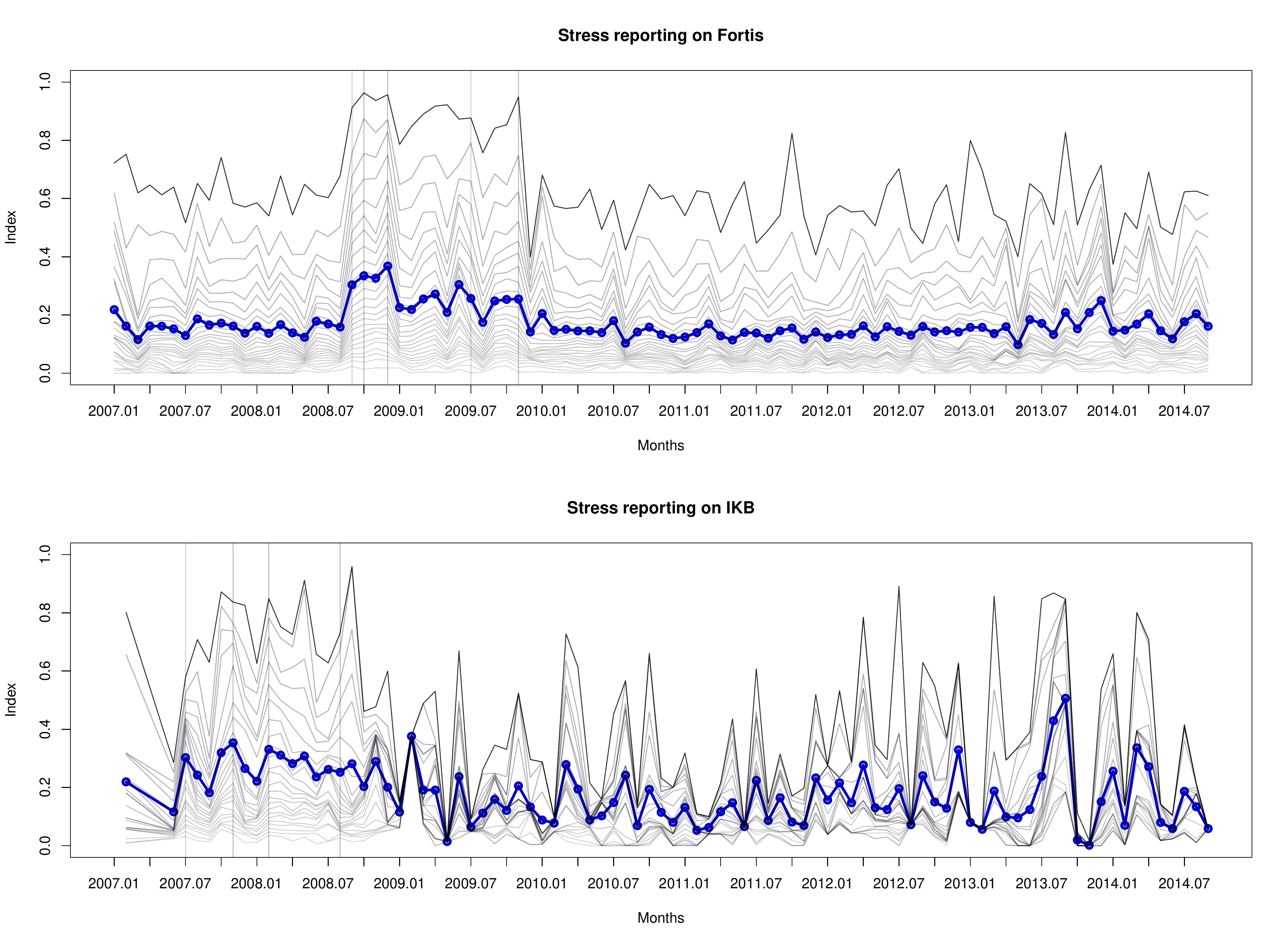}

\protect\caption{Indices (blue) for banks Fortis and IKB indicating the levels of bank stress-related reporting, with faded lines showing every 4th percentile up to the 98th. Vertical lines indicate recorded events.}

\label{fig_single_banks}
\end{figure*}

One of the early failures among European financial institutions occurred to the Benelux-based Fortis. As was also highlighted in the above described top excerpts for Belgium, Fortis and the rescue procedure was at the core of the discussion as the crisis erupted. We focus on the evolution of the distress index for Fortis, as is shown in Fig. \ref{fig_single_banks}.
To start with, we can observe that elevated values for the stress index coincide with distress events.

By the first event in September 2008, the index rises to 0.30, which marks the start of a prolonged period of elevated stress. The top-ranked excerpts relate to a range of different issues, such as worries about lacking confidence in the markets and the systemic nature of the unfolding crisis:

\begin{quote}\end{quote}
\begin{quote}\emph{
"Jean-Claude Juncker, also the prime minister of Luxembourg, was asked whether the part nationalisation 
of Dutch-Belgian bank Fortis FOR. -BR and a new injection of liquidity into money markets by the European
Central Bank would restore market confidence. ``I can only hope that confidence will come back --
financial markets should not forget to take a close look at the health of fundamental data of several
banks -- and that this casino game, that's going on independently from the good fundamentals, stops,''
he told reporters on the sidelines of a meeting in parliament. Belgian, Dutch and Luxembourg
governments rescued Fortis over the weekend to prevent a domino-like spread of failure by buying its
shares for 11.2 billion euros."}

(Reuters 2008-09-29, 
relevance 0.963, rank 1)
\end{quote}

\begin{quote}\emph{"Investors also worried if a proposed U.S. rescue would stem the contagion that pushed the British government to takeover troubled mortgage lender Bradford \& Bingley BB.L and three European governments to partially nationalize banking and insurance group Fortis FOR.BRFOR.AS."}

(Reuters 2008-09-29, 
relevance 0.923, rank 6)
\end{quote}

In October 2008, the top excerpts discuss the continuing developments such as the Benelux governments "carving up" Fortis to sell to private entities, including French BNP Paribas buying control of the arms in Belgium and Luxembourg. Further excerpts highlight the cross-border aspect of the interventions, and the issues it entails:

\begin{quote}\emph{
"The Fortis deal is the biggest cross-border rescue since the full force of the credit crisis swept across the Atlantic into Europe last month, upending banks and rattling saver confidence."} 

(Reuters 2008-10-06, 
relevance 0.945, rank 7)
\end{quote}
\begin{quote}\emph{
"Dutch Finance Minister Wouter Bos fanned Belgian resentment by telling journalists: `Many of the problems were hidden in the Belgian part of the Fortis group.''"} 

(Reuters 2008-10-05, 
relevance 0.945, rank 8)
\end{quote}

This repeats the message of the already cited news for the UK in September 2008, that "there is no consensus on who would bail out a failed multinational bank", highlighting how the use of text descriptions can provide deeper insight into the multifaceted developments underlying a model signal.





Without a detailed analysis of the discussion around the IKB Bank, we can again conclude from Fig. \ref{fig_single_banks} that the stress index takes high values during the realized events. Generally, the top-ranked discussion herein correlates to a large extent with the early top-ranked discussion for Germany, as was above exemplified. The discussion around the distress events relates to early indications of stress, ties to other German banks and government actions taken during and after the stress episodes. After a period of elevated stress during 2007--2008, the figure illustrates that stress is still fairly volatile and that the most extreme percentiles still take large values. This may relate to the fact that discussion keeps relating to the 2007--2008 distress events, in that the solution to the stress events was an acquisition by a investment company. The private equity firm Lone Star acquired IKB Bank in 2008 with the aim of restructuring and selling the bank, and accordingly any rumours still link it to the original stress discussion during the global financial crisis. Such references to past major stress events may however also be an indication of current concerns about financial stress, thus worth signaling in order to allow further investigation.

\section{Conclusions}

We have presented a deep-learning-based approach that combines two types of data, news text and basic event information, with the aim of linking the two to describe observed and predicted events. The approach entails unsupervised learning on text in order to model its language and provide semantic vector representations that are used as features for predictive modeling of events. The neural-network-based method that we put forward is able to work with a very small set of events, matched with text through a heuristic, in order to discern what type of language and passages in the text are actually relevant to the modeled event type and phenomenon. The semantic modeling utilizes large amounts of text data to infer abstractions that counter the high variability and sparsity of language, thus supporting prediction of infrequent events. 

The semantic-predictive model can produce indices that indicate the level of relevant discussion over time, overall or related to specific entities or groups thereof. The indices can highlight interesting patterns and offer guidance in the search for relevant events, whereas the model very directly provides means to rank and retrieve pieces of text from news articles in order to describe the quantitative signal.

We demonstrate the usefulness of the method and the possibilities of the approach in general within the study of financial risk, by modeling bank distress events. The indices reflect the level of current reporting related to bank stress over time at multiple levels: for Europe in general, for individual countries and for specific banks. Guided by the indices, users may focus their search and retrieve the relevant reporting of the time, in order to understand the developments regarding, in this case, government interventions and rescues. Our quantitative evaluation of the stress index shows good results and provides an important quality assurance of the descriptions.

The method and our analysis exemplify how text may offer an important complementary source of information for financial and systemic risk analytics, which is readily available, current and rich in descriptive detail. In contrast to traditional information sources, text data offers a possible route to circumvent the issues of privileged access, lagging publication and low granularity, but most importantly does it very directly offer value through the explanatory power of the event-related human language descriptions accompanying the plane signal. We expect the method to be also directly applicable to describe events beyond the financial domain, relating to geopolitics and other significant topics.

We recognize that deep learning approaches are useful in particular to handle the complexities of such new types of data, while offering necessary flexibility when exploring new fields of analysis. Seeking to harness the expressiveness of text, we should continue to look to computational linguistics for support in terms of theoretical foundations and tools.

While we show that it is possible to predict relevance and retrieve informative descriptions of events, we merely scratch the surface of the vast text material in any given cross section with our current method of presentation. A challenge remains in developing methods that are able to meaningfully summarize the broader base that may include a long tail of weakly signaling, subtle expressions. Such signals may be particularly important in order to register and track developments before they materialize in severe and obvious events. Likewise, to really make use of text data as a complement rather than a replacement, traditional sources and text should be integrated in a unified modeling framework in order to achieve the best predictive performance possible, while also keeping the opportunity to explore the descriptions to that signal open.

\section{Appendix}





\begin{figure*}
\includegraphics[width=0.99\textwidth]{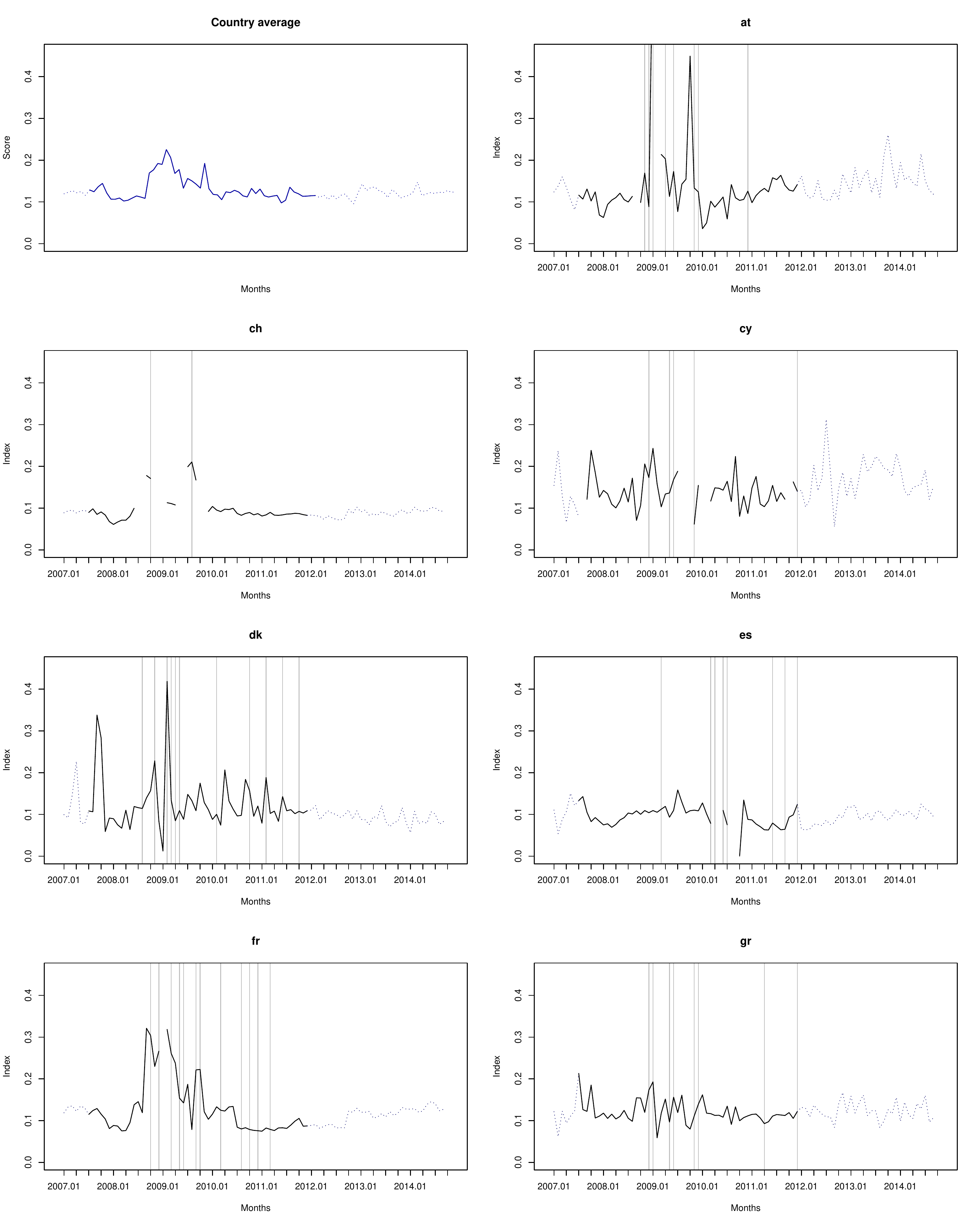}
\protect\caption{Distress index for Austria, Switzerland, Cyprus, Denmark, Spain, France, Greece and average of all modeled countries. Vertical lines indicate bank-level distress events and dotted lines out-of-sample predictions.}

\label{fig_all_index1}
\end{figure*}

\begin{figure*}
\includegraphics[width=0.99\textwidth]{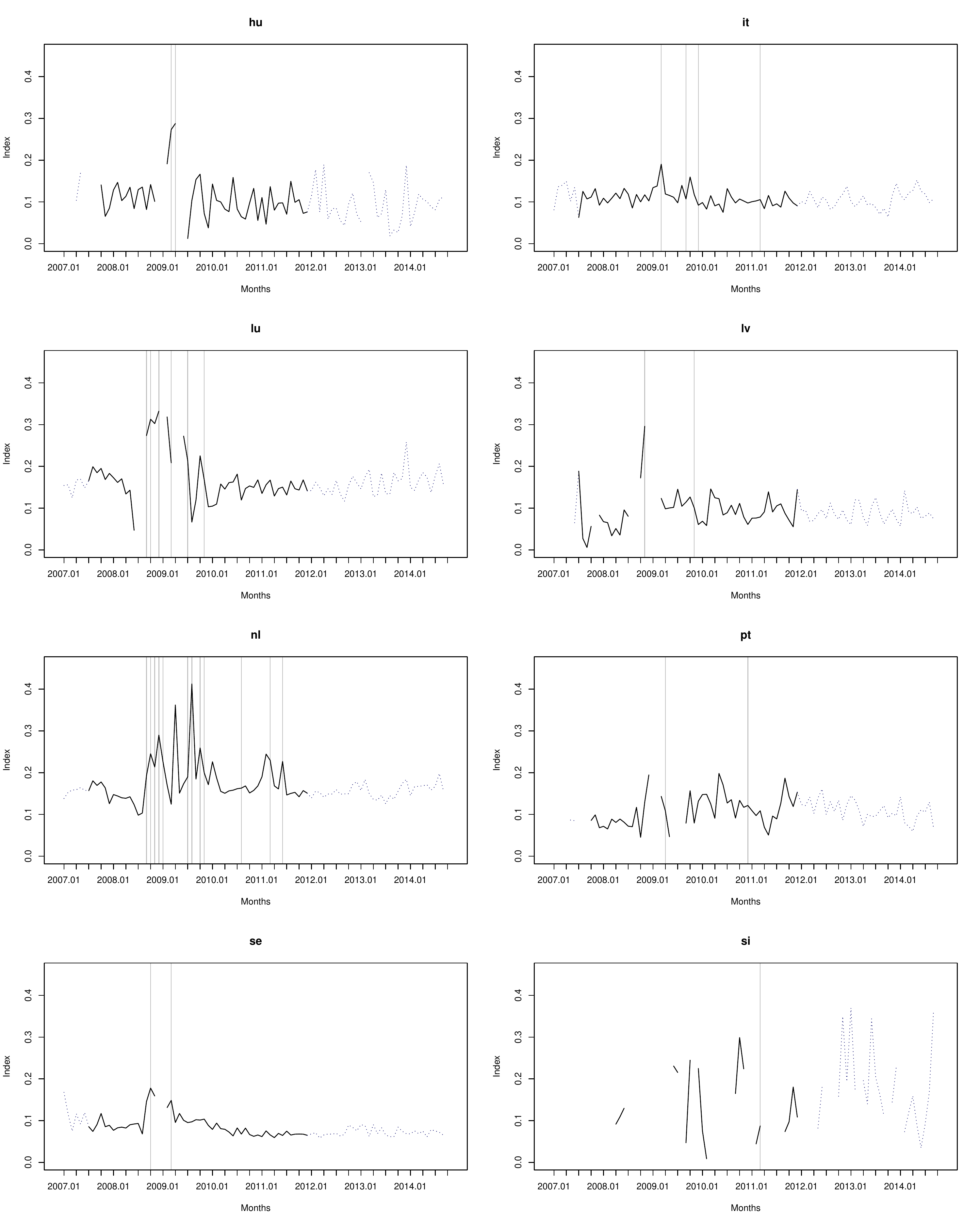}
\protect\caption{Distress index for Hungary, Italy, Luxembourg, Latvia, Netherlands, Portugal, Sweden and Slovenia. Vertical lines indicate bank-level distress events and dotted lines out-of-sample predictions.}

\label{fig_all_index2}
\end{figure*}

Figs. \ref{fig_all_index1} and \ref{fig_all_index2} provide country-level indices for the countries not included in Figs. \ref{fig_excerpts1} and \ref{fig_excerpts2}, as well as the non-weighted average of all country indices. The individual banks and countries they are mapped to are listed in Table \ref{bank_table}.

\begin{table}
\centering
\scriptsize
\begin{tabular}{llll}
Bank & Country & Bank & Country\tabularnewline
\hline
ABN Amro & NL & Fionia (Nova Bank) & DK \tabularnewline
ATE Bank & GR & First Business Bank & GR \tabularnewline
Aareal Bank & DE & Fjordbank Mors & DK \tabularnewline
Aegon & NL & Fortis Bank & LU, NL, BE \tabularnewline
Agricultural Bank of Greece & GR & HBOS & UK \tabularnewline
Allied Irish Banks & IE & HSH Nordbank & DE \tabularnewline
Alpha Bank & GR & Hellenic & GR \tabularnewline
Amagerbanken & DK & Hypo Alpe Adria Group & AT \tabularnewline
Anglo Irish Bank & IE & Hypo Real Estate & DE \tabularnewline
Attica Bank & GR & Hypo Tirol Bank & AT \tabularnewline
BBK & ES & IKB & DE \tabularnewline
BNP Paribas & FR & ING & NL \tabularnewline
BPCE & FR & Irish Life \& Permanent & IE \tabularnewline
BPP & PT & Irish Nationwide Building Society & IE \tabularnewline
Banca Civica & ES & KBC & BE \tabularnewline
Banca Popolare & IT & Kommunalkredit & AT \tabularnewline
Banca Popolare di Milano & IT & LBBW & DE \tabularnewline
Banco Mare Nostrum & ES & Lloyds TSB & UK \tabularnewline
Banco Popolare & IT & Lokken & DK \tabularnewline
Banco de Valencia & ES & Magyar Fejlesztesi Bank Zrt & HU \tabularnewline
Bank of Cyprus & CY & Marfin Popular Bank & CY \tabularnewline
Bank of Ireland & IE & Max Bank & DK \tabularnewline
Bankia & ES & Monte dei Paschi di Siena & IT \tabularnewline
Banque Populaire & FR & Mortgage and Bank of Latvia & LV \tabularnewline
Bawag & AT & National Bank of Greece & GR \tabularnewline
BayernLB & DE & NordLB & DE \tabularnewline
CAM & ES & Nordea & SE \tabularnewline
Caisse d'Epargne & FR & Northern Rock & UK \tabularnewline
Caixa General de Depositos & PT & Nova Ljubljanska banka & SI \tabularnewline
Caja Castilla-La Mancha & ES & Novacaixagalicia & ES \tabularnewline
Caja Espana & ES & OTP Bank Nyrt & HU \tabularnewline
Carnegie Investment Bank & SE & Panellinia Bank & GR \tabularnewline
Catalunyacaica & ES & Pantebrevsselskabet & DK \tabularnewline
Commerzbank & DE & Parex & LV \tabularnewline
Cooperative Central Bank & CY & Piraeus Bank & GR, CY \tabularnewline
Credit Agricole & FR & Proton Bank & GR \tabularnewline
Credit Mutuel & FR & RBS & UK \tabularnewline
Credito Valtellinese & IT & RZB Group & AT \tabularnewline
Cyprus Development Bank & CY & Roskilde Bank & DK \tabularnewline
Cyprus Popular & CY & SNS Reaal & NL \tabularnewline
Danske Bank & DK & SachsenLB & DE \tabularnewline
Dexia & BE, FR, LU & Societe Generale & FR \tabularnewline
Dunfermline & UK & Swedbank & SE \tabularnewline
EBH & DK & T-Bank & GR \tabularnewline
EBS Building Society & IE & UBS & CH \tabularnewline
EFG Eurobank & GR & UNNIM & ES \tabularnewline
Eik Bank & DK & USB Bank & CY \tabularnewline
Erste Bank & AT & VBAG & AT \tabularnewline
Ethias & BE & Vestjysk & DK \tabularnewline
FHB Jelzalogbank Nyrt & HU & WestLB & DE \tabularnewline
Finansieringsselskabet & DK \tabularnewline

\tabularnewline
\end{tabular}
\\
\protect\caption{Target banks and their countries.}
\label{bank_table}
\end{table}

\section*{Acknowledgment}

The research has been funded by the Graduate School of \AA bo Akademi University and the Turku Centre for Computer Science Graduate Programme.
The authors are grateful to Filip Ginter, J\'ozsef Mezei, Tuomas Peltonen and Niko Schenk for their helpful comments. The paper also has benefited from presentation at 
the Finnish Economic Association XXXVII Annual Meeting (KT-p\"aivat), 12 February 2015, in Helsinki, Finland; the RiskLab/Bank of Finland/European Systemic Risk Board (ESRB) Conference on Systemic Risk Analytics (SRA), 24 September 2015, in Helsinki; the workshop of GI-FG Neuronale Netze and German Neural Networks Society, New Challenges in Neural Computation (NC\^{}2), 10 October 2015, in Aachen, Germany; the IEEE Conference on Computational Intelligence in Financial Engineering and Economics (CIFEr), 9 December 2015, in Cape Town, South Africa; the Financial Stability Seminar at the Riksbank, 12 January 2016, in Stockholm, Sweden; and the 23rd Annual Conference of the Multinational Finance Society, 26 June 2016, in Stockholm, Sweden; as well as from featuring on Bloomberg View. The work presented in this paper has been replicated by Thomson Reuters.



%


\newpage

\bibliographystyle{plain}
\bibliography{references}


\end{document}